\documentclass{article}
\usepackage[utf8]{inputenc}

\title{Multimodal Transformer-based Model for Buchwald-Hartwig and Suzuki-Miyaura Reaction Yield Prediction}

\usepackage{authblk}
\usepackage[numbers]{natbib}
\usepackage{graphicx}
\usepackage{tabularx}
\usepackage{multicol}
\usepackage{makecell}
\usepackage{multirow}
\usepackage{booktabs}
\usepackage{caption}
\usepackage{subcaption}

\usepackage[a4paper, total={5.5in, 9in}, margin=1.5in]{geometry}

\author[1]{Shimaa Baraka }
\author[2, 3]{Ahmed M. El Kerdawy}
\date{}

\affil[1]{Integrant, Inc., Cairo, Egypt}
\affil[2]{Department of Pharmaceutical Chemistry, Faculty of Pharmacy, Cairo University, Cairo, Egypt}
\affil[3]{Department of Organic and Pharmaceutical Chemistry, School of Pharmacy, Newgiza University (NGU), Newgiza, Cairo, Egypt}

\begin{document}
\maketitle

\begin{abstract}
  Predicting the yield percentage of a chemical reaction is useful in many aspects such as reducing wet-lab experimentation by giving the priority to the reactions with a high predicted yield. In this work we investigated the use of multiple type inputs to predict chemical reaction yield. We used simplified molecular-input line-entry system (SMILES) as well as calculated chemical descriptors as model inputs. The model consists of a pre-trained bidirectional transformer-based encoder (BERT) and a multi-layer perceptron (MLP) with a regression head to predict the yield. We experimented on two high throughput experimentation (HTE) datasets for Buchwald–Hartwig and Suzuki–Miyaura reactions. The experiments show improvements in the prediction on both datasets compared to systems using only SMILES or chemical descriptors as input. We also tested the model's performance on out-of-sample dataset splits of Buchwald–Hartwig and achieved comparable results with the state-of-the-art.

In addition to predicting the yield, we demonstrated the model's ability to suggest the optimum (highest yield) reaction conditions.
The model was able to suggest conditions that achieves 94\% of the optimum reported yields. This proves the model to be useful in achieving the best results in the wet lab without expensive experimentation.
\end{abstract}

\section{Introduction}
Predicting the yield percentage of a chemical reaction is an important problem in drug discovery to guide chemists to the optimal conditions to achieve the highest yield. It also decreases wet-lab experimentation by focusing only on the highest yield reactions. In this work we experimented with using multi-modal input to predict the yield and suggest the optimal conditions of Buchwald–Hartwig and Suzuki–Miyaura reactions as an example to widely utilized reaction types. 


\section{System}
  Machine learning was proven to be a useful tool in predicting reaction yield. Previous work depended on one form of descriptors over the other. For example, reference \citet{yieldSMILES} used SMILES annotations and language-based deep learning models to predict the yield. Whereas, reference \citet{struct} used structure-based fingerprints. Reference \citet{yieldDesc} experimented with multiple models, such as Random Forest, using chemical descriptors as an input. To allow our model to benefit from various types of inputs, we used both SMILES annotation and chemical descriptors as inputs. Chemical descriptors directly encode compound properties that can be beneficial in predicting the yield. On the other hand, SMILES representation encodes the syntax of compound structure and it is much easier to have a pre-trained model on large corpus for its representation. 
  
\begin{center}
    \begin{figure}[!ht] \includegraphics[scale=0.3]{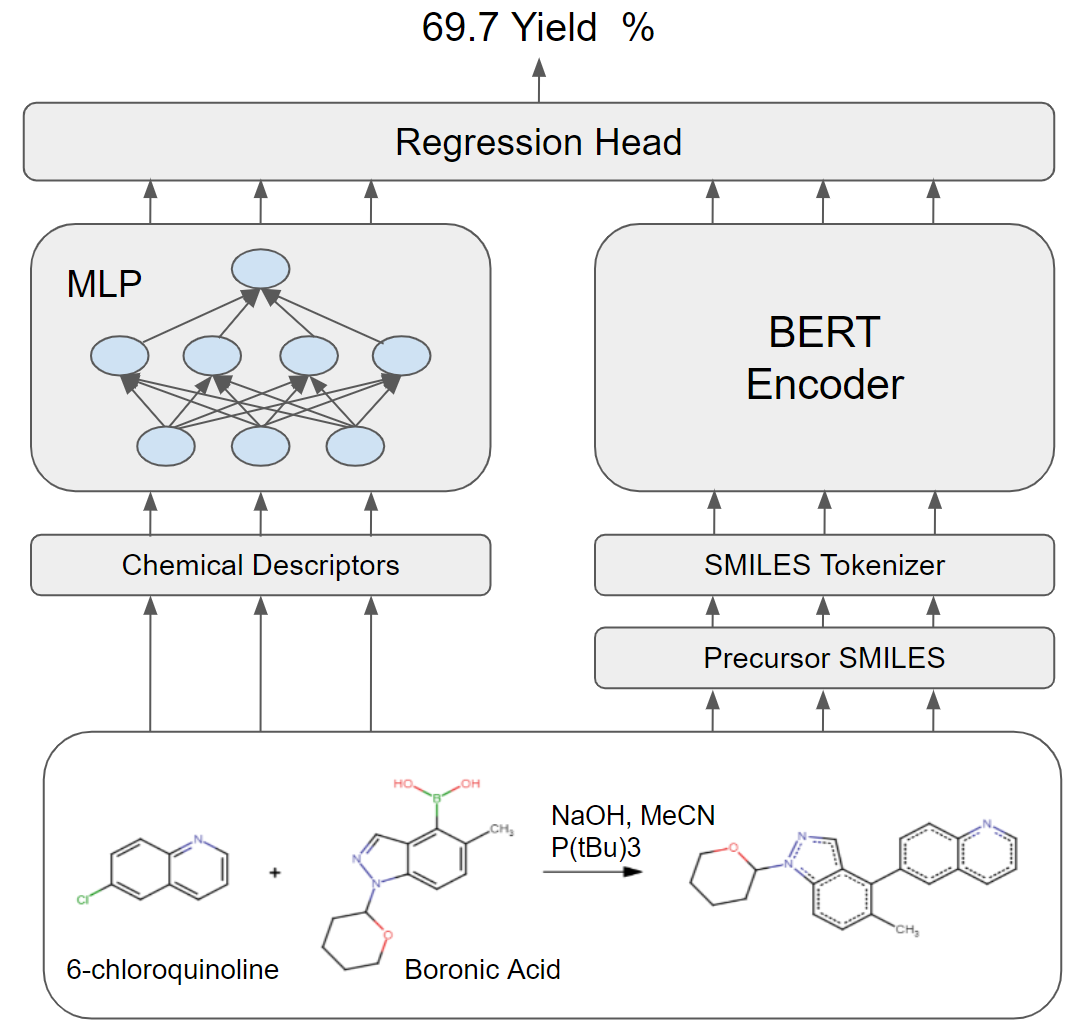}
      \centering
      \caption{An example of a reaction forward path.}
    \label{fig:model}
    \end{figure}
    \end{center}

  We developed a model with two input channels: a bidirectional transformer-based encoder (BERT) \cite{bert} for the SMILES representation and a multi-layer perceptron (MLP) for the chemical descriptors. The two processed inputs are concatenated and followed by a regression layer to predict the yield. For the SMILES encoder we used the pre-trained transformer-based model obtained from ref. \citet{lang_model} and fine-tuned it along with the MLP for every dataset. Figure \ref{fig:model} illustrates the model flow for an example reaction.
\section{Data}
    We experimented with two HTE datasets: One for Buchwald–Hartwig reaction \cite{yieldDesc} and the other for Suzuki–Miyaura reaction \cite{suzuki}. Buchwald–Hartwig's dataset has 3955 data points that include the combination of 15 Aryl halides, 4 ligands and 3 bases. The dataset also takes into account the inhibitory effect of 23 different isoxazole additives. As for Suzuki–Miyaura dataset, a balanced dataset of 15 pairs of electrophiles and nucleophiles are used with various ligand, reagent and solvent combinations resulting in 5760 data points.

\section{Results}
  We split both datasets by 70/30 ratio to be able to compare the results with that of the previously reported literature. The reported results is using 10 random folds of the datasets. Hyper-parameter optimization was carried on 1/7 of the training set of the first fold, similar to ref. \citet{yieldSMILES}. Table \ref{tab:bw_results} and \ref{tab:sm_results} show the results on the used datasets.
   \begin{table}[!ht]
  \captionsetup{justification=justified, singlelinecheck=false}
   \caption{Results on Buchwald–Hartwig reaction.}
      \centering
      \begin{tabularx}{\textwidth}{X c c}
        \toprule
         \textbf{Buchwald–Hartwig}  & RMSE & $\mathrm{R^2}$ \\
        \midrule
        one-hot \cite{commentyieldDesc} & - & 0.89\\ 
        DFT \cite{yieldDesc} & 7.8 & 0.92\\
        MFF \cite{struct} & - & 0.927 ± 0.007\\
        BERT \cite{yieldSMILES} & - & 0.951 ± 0.005  \\
        Multimodal BERT (proposed model) & 5.5 ± 0.3 & \textbf{0.959 ± 0.005} \\
        \bottomrule
      \end{tabularx}
      
     \label{tab:bw_results}   
    \end{table}

    \begin{table}[!ht]
    \captionsetup{justification=justified, singlelinecheck=false}
    \caption{Results on Suzuki–Miyaura reaction.}
      \centering
      \begin{tabularx}{\textwidth}{X c c}
        \toprule
         \textbf{Suzuki–Miyaura} & RMSE & $\mathrm{R^2}$ \\
        \midrule
        BERT \cite{yieldSMILES} & - &  0.81 ± 0.01  \\
        Multimodal BERT (proposed model)& 11.5 ± 0.3 & \textbf{0.833 ± 0.01} \\
        \bottomrule
      \end{tabularx}
      
      \label{tab:sm_results}
    \end{table}

\section{Out-of-sample Prediction}
  To test the performance of the model on out-of-sample data, we used specially split dataset for the Buchwald–Hartwig reaction so that isoxazole additives in the test data are different from that of the training data. We tested on the four splits provided by ref. \citet{yieldDesc}. As shown in table \ref{tab:bw_of_of_sample} the model showed promising performance with $\mathrm{R^2}$ of 0.72, despite the shift, which is comparable to the results in ref. \citet{yieldSMILES}.
  
  To test the distribution shift on Suzuki–Miyaura's dataset we split the reaction space so that the ligands in the training set are different from that of the test set. We tested 4 different splits, each with 1/4 of the available ligands in the test set only. The performance dropped to 0.2±0.25 $\mathrm{R^2}$ and 25±6 RMSE. These results indicate a much bigger difference between the training and the test sets distributions in Suzuki–Miyaura's dataset relative to Buchwald–Hartwig's dataset. 
  
     \begin{table}[!ht]
     \captionsetup{justification=justified, singlelinecheck=false}
      \caption{Out-of-sample results on Buchwald–Hartwig reaction.}
      \centering
      \begin{tabularx}{\textwidth}{X c c c c c c}
        \toprule
         Model & Test 1 & Test 2 & Test 3 & Test 4  &
         \makecell{Average \\ $\mathrm{R^2}$} & \makecell{Average \\ RMSE} \\
        \midrule
        DFT \cite{yieldDesc} &  0.8 & 0.77 & 0.64 & 0.54 & 0.69 & -\\
        one-hot \cite{commentyieldDesc} & 0.69 & 0.67 & 0.49 & 0.49 & 0.59 & -\\
        MFF \cite{struct} & 0.85 & 0.71 & 0.64 &  0.18  & 0.60 & - \\
        BERT \cite{yieldSMILES} & 0.84 & 0.84 & 0.75 & 0.49 & 0.73 & - \\
        Multimodal BERT \\(proposed model) &0.86 & 0.86 & 0.62 &  0.54& 0.72 & 13.8 ± 3.6\\
        \bottomrule
      \end{tabularx}
      \label{tab:bw_of_of_sample}   
    \end{table}

\section{Selecting Optimum Reaction Conditions}
  One possible usage of predicting yield is selecting conditions that lead to the highest yield without wet-lab experiments. In this experiment we focused on Suzuki–Miyaura dataset as Buchwald–Hartwig's dataset contains isoxazole additives, which is not a natural reaction condition. To test the ability of the system to suggest the optimum conditions, we compared the yield of the top conditions suggested by the model to the top reported yield for every pair of reactants in Suzuki–Miyaura \cite{suzuki} test dataset. The best reported conditions of the 15 pairs give 0.87 average yield while the conditions suggested by the model gives 0.82 average yield. This means the suggested yield achieves about \textbf{94\%} of the reported optimal yield. A random selection of conditions without any model gives 0.43 average yield, which is about 50\% of the optimal yield. This proves the efficiency of the model in suggesting the reaction conditions required for achieving the highest yield. Table \ref{tab:example_cond} shows the top 5 suggested conditions for the highest yield for reacting 6-Bromoquinoline with Boronic Acid.
  
  To check if the exact reported conditions are suggested, we calculated the accuracy of the existence of the reported optimal condition in the top-k\% suggested conditions. Table \ref{tab:acc_cond} shows accuracy per top-k\%. We can see from table \ref{tab:example_cond} that the difference between the top experimental entries can be quite small, thus judging the model by the mean estimated yield, as above, is better than the accuracy of predicting the actual conditions. 
  
  \begin{table}[ht!]
  \captionsetup{justification=justified, singlelinecheck=false}
  \caption{Suggested conditions (a) for reacting 6-Bromoquinoline and Boronic Acid Vs. experimental best combination (b).}
      \centering
     \captionsetup{justification=centering}
      \subcaption{Suggested conditions}
      \begin{tabular}{c c c c c}
        \toprule
         Liagnd & Reagent & Solvent & \makecell{Estimated \\ Yield}  & \makecell{Actual \\ Yield} \\
        \midrule
        CatacXium A & LiOtBu & MeCN & 1 & 0.959  \\
        P(tBu)3 & NaHCO3 & THF & 0.999 & 0.940 \\
        P(o-Tol)3 & LiOtBu & MeOH & 0.970 &0.881\\
        CatacXium A & NaHCO3 & THF & 0.968 & 0.969  \\
        P(Ph)3 & K3PO4 & MeOH & 0.968 &0.947\\
        \bottomrule
      \end{tabular}
      \centering
      \captionsetup{justification=centering}
      \subcaption{Actual conditions}
      \begin{tabular}{c c c c }
          \toprule
          Liagnd & Reagent & Solvent & \makecell{Actual \\ Yield} \\
          \midrule
          CatacXium A & NaHCO3 & THF & 0.969\\
          P(Ph)3 & LiOtBu & MeCN & 0.968 \\
          CatacXium A & LiOtBu & MeCN & 0.959 \\
          CatacXium A & KOH & MeCN & 0.956\\
          P(Ph)3 & K3PO4 & MeOH & 0.947\\
      \end{tabular}
    \label{tab:example_cond}   
    \end{table}

  \begin{table}[!ht]
  \caption{Accuracy of top-k percent of suggested conditions.}
      \centering
      \begin{tabular}{c   c}
        \toprule
         Top-k Percent & Accuracy \\
        \midrule
        5\% & 53.3 \\
        10\% & 73.3 \\
        15\% & 80 \\
        30\% & 86.6 \\
        \bottomrule
      \end{tabular}
      \label{tab:acc_cond} 
    \end{table}

\section{Conclusion}
Using multiple modalities in predicting the yield of a reaction condition proves to increase the quality of the prediction. Moreover, it is capable of suggesting condition combinations for achieving high reaction yield. This proves the model to be useful in guiding the chemist in the process of selecting the optimal conditions.  
\bibliographystyle{plainnat}
\bibliography{poster}

\end{document}